# Improving VisNet for Object Recognition

Mehdi Fatan Serj [1,*], C. Alejandro Parraga [1], Xavier Otazu [1]
[1]Computer Vision Centre, Autonomous Univiversity of Barcelona, Spain
mfatan@cvc.uab.cat

**Abstract**

Object recognition plays a fundamental role in how biological organisms perceive and interact with their environment. While the human visual system performs this task with remarkable efficiency, reproducing similar capabilities in artificial systems remains challenging. This study investigates *VisNet*, a biologically inspired neural network model, and several enhanced variants incorporating radial basis function neurons, Mahalanobis distance–based learning, and retinal-like preprocessing for both general object recognition and symmetry classification. By leveraging principles of Hebbian learning and temporal continuity—associating temporally adjacent views to build invariant representations—VisNet and its extensions capture robust and transformation-invariant features. Experimental results across multiple datasets, including MNIST, CIFAR-10, and custom symmetric object sets, show that these enhanced VisNet variants substantially improve recognition accuracy compared with the baseline model. These findings underscore the adaptability and biological relevance of VisNet-inspired architectures, offering a powerful and interpretable framework for visual recognition in both neuroscience and artificial intelligence.

## 1 Introduction

Artificial Intelligence (AI) has experienced extraordinary progress in recent decades, much of which has been driven by innovations inspired by neuroscience. These approaches range from broadly inspired frameworks that borrow conceptual principles to biologically plausible models that closely mimic neural mechanisms and architecture. Biologically inspired methods—such as Convolutional Neural Networks (CNNs)—have revolutionized computer vision, enabling human-level performance in tasks such as object recognition, scene understanding, and visual classification (Krizhevsky et al., 2012; LeCun et al., 2015; DiCarlo et al., 2012a; Serre et al., 2007). This rapid progress has been influenced by insights from biological vision, where convolutional operators were originally motivated by the receptive fields of neurons in the early visual cortex (Fukushima, 1980; Hubel and Wiesel, 1962a). Recent developments, including hierarchical processing, attention mechanisms, and predictive coding, further demonstrate how neural principles continue to shape modern AI models (Kietzmann et al., 2019). Despite these successes, most AI systems still operate as opaque "black boxes," offering little insight into their internal representations (Lipton, 2018). Their limited transparency and interpretability hinder broader adoption in safety-critical applications, such as healthcare, robotics, and autonomous navigation. As a result, a growing research direction seeks to develop computational architectures that are not only powerful but also biologically plausible and interpretable. Such models offer two notable advantages: (1) they provide insight into perceptual mechanisms in the brain through interpretable internal representations, and (2) they generate hypotheses about human cognition that can be empirically tested (Kriegeskorte and Douglas, 2018; Richards et al., 2019). One such model is *VisNet*, a four-layer unsupervised neural network introduced by Rolls and colleagues (Wallis and Rolls, 1997a; Rolls and Stringer, 2006), designed to reproduce hierarchical visual processing in the primate visual cortex. The model relies on Hebbian learning (Hebb, 1949a) and a temporal trace rule to associate temporally adjacent views of the same object, allowing it to form invariant object representations under transformations such as rotation and scaling (Rolls and Stringer, 2006; Wallis and Rolls, 1997a). Unlike conventional architectures that focus primarily on spatial features, VisNet's capacity to learn from temporal input sequences facilitates dynamic object recognition (Hochreiter and Schmidhuber, 1997). This property parallels the way the human brain learns to recognize objects across variable viewing conditions—such as changes in angle, scale, and illumination (DiCarlo and Cox, 2007). By incrementally constructing invariant representations, VisNet provides a transparent and interpretable framework for understanding visual processing while offering strong potential for computational applications. An especially compelling aspect of visual perception is symmetry, which plays a central role in how both humans and animals recognize and categorize objects. Human observers can often identify three-dimensional symmetric objects from a single view, even one not aligned with the symmetry plane (Vetter et al., 1994). Symmetry perception thus provides efficient cues for recognition but presents considerable computational difficulty for artificial systems. Challenges include the arbitrary orientation of symmetric patterns, the interplay of reflectional and rotational symmetries, and the complexities introduced by transformations during data augmentation (Funk and Liu, 2016; Zabrodsky and Weinshall, 1992; Liu et al., 2010; Seo et al., 2022). Understanding these challenges is essential, as symmetry detection lies at the intersection of neuroscience and computer vision, with implications for artificial intelligence, robotics, and biological vision research. The objective of this work is to evaluate VisNet's effectiveness in classifying and recognizing symmetric objects. Given its biologically grounded mechanisms for developing invariant representations, VisNet offers a unique computational basis for exploring the relationship between symmetry, temporal learning, and visual invariance (Fukushima, 1980; Friston, 2005). The insights derived from this study contribute to advancing biologically inspired models of perception and bring us closer to building interpretable AI systems that integrate the principles of neuroscience with modern computational vision (Krizhevsky et al., 2012).

# 2 Related Work

## 2.1 Computational Models of Vision

Computational models of vision have evolved significantly over the past few decades, beginning with the foundational theoretical work of David Marr (Marr, 1982), who established a framework for understanding early visual processes such as edge detection, stereo vision, and motion perception. Building on these principles, Fukushima introduced the *Neocognitron* (Fukushima, 1980), a hierarchical architecture inspired by the simple and complex cells described by Hubel and Wiesel (Hubel and Wiesel, 1962b). The Neocognitron demonstrated how layered feature extraction could support object recognition, laying the conceptual groundwork for modern deep learning models. By the 1990s, computational models increasingly incorporated neurophysiological evidence. Daly's *Visual Difference Predictor* (Daly, 1993) modeled perceptual visibility using human contrast sensitivity, while Riesenhuber and Poggio's *HMAX* model (Riesenhuber and Poggio, 1999) captured selectivity and invariance mechanisms analogous to those observed in the primate ventral visual stream. In parallel, Olshausen and Field (Olshausen and Field, 1996) proposed *sparse coding* models, demonstrating how cortical representations of natural images can be formed from a limited set of basis functions similar to receptive fields in V1. Around the same period, Rolls introduced the *VisNet* architecture (Rolls et al., 1997; Wallis and Rolls, 1997a), a self-organizing hierarchical network that learned transformation-invariant object representations through biologically plausible mechanisms such as Hebbian and trace learning. In addition, predictive coding frameworks (Rao and Ballard, 1999) argued that the brain integrates vision through top-down predictions and bottom-up error correction—a concept now central to computational neuroscience. Serre et al. (Serre et al., 2007) later proposed *dynamic routing networks*, which combined feedforward and feedback information, further improving biological plausibility. In recent years, deep neural networks have incorporated many of these biologically inspired principles. Convolutional Neural Networks (CNNs) (Yann LeCun, 1998) introduced hierarchical feature extraction reminiscent of the visual cortex, while subsequent advancements such as *AlexNet* (Krizhevsky et al., 2012), *VGG* (Karen Simonyan, 2015), and *ResNet* (Kaiming He, 2016) achieved unprecedented performance on large-scale visual recognition benchmarks. More recently, Vision Transformers (*ViTs*) (Dosovitskiy et al., 2021) have extended this paradigm by leveraging self-attention mechanisms to capture long-range dependencies across the entire visual field, aligning conceptually with the brain's ability to integrate spatially distributed information. From Marr's early theoretical models to contemporary biologically inspired and biologically plausible architectures, computational vision research has progressively integrated hierarchical processing, predictive learning, and efficient coding principles. These developments continue to narrow the gap between artificial systems and the complexity of human visual perception. Consistent with this trajectory, the present study focuses exclusively on biologically plausible learning mechanisms as a means to develop interpretable and robust models of object recognition.

## 2.2 Symmetry Detection and Recognition

Symmetry is a defining property of many ecologically significant objects, including fruits, leaves, and animal bodies (Thompson and Bonner, 1992). Across the animal kingdom, where distinguishing allies from predators is essential, symmetry perception plays a crucial role in survival (Troscianko et al., 2009). In humans, symmetry is strongly linked with perceptions of balance, health, and aesthetic appeal (Treder, 2010). In other species, such as birds, symmetry contributes to behaviors like mate selection, where it often serves as an indicator of genetic quality (Gamble and Wright, 2010). Despite its clear behavioral and perceptual importance, the neural and computational mechanisms underlying symmetry detection remain only partially understood. Functional MRI (fMRI) studies have identified that symmetric patterns preferentially activate specific higher-level regions of the visual cortex, including extrastriate areas involved in spatial integration (Sasaki et al., 2005). Psychophysical research further highlights the influence of early, low-level processes on symmetry perception, suggesting a tight interaction between bottom-up and top-down visual mechanisms (Treder, 2010). From a neurocomputational perspective, early symmetry detection approaches focused on pairing symmetric features (Loy and Eklundh, 2006; Rainville and Kingdom, 2000) or locating symmetry axes (Osorio, 1996; Akbarinia et al., 2017; Parraga et al., 2019) using low-level operators similar to Gabor filters. These models, however, rarely addressed higher-order hierarchical integration. From an engineering standpoint, symmetry detection techniques have progressed from geometric rule-based methods—such as reflection axis estimation (Liu and Xie, 2010)—to modern deep learning systems capable of recognizing symmetry in complex and cluttered visual scenes (Brachmann and Redies, 2016). More recently, Wu (Wu and Liu, 2022) introduced a convolutional neural network specifically designed to assess both reflectional and rotational symmetries, marking a step toward bridging biological and machine-based symmetry recognition. Nevertheless, symmetry recognition remains a challenging computational problem. A perceptual skill that arises effortlessly in humans continues to confound artificial systems. This disparity has even led researchers to propose symmetry-based tests as robust visual "CAPTCHAs" resistant to machine decoding (Funk and Liu, 2016). These challenges underscore the need for models—such as VisNet—that leverage biologically plausible learning mechanisms to approach the human brain's remarkable efficiency in recognizing and reasoning about symmetrical structures. In this paper, we build on these insights by extending the VisNet framework and empirically examining its ability to learn invariant, symmetry-sensitive representations using biologically plausible mechanisms across a range of visual tasks.

# 3 Background: VisNet

VisNet (Wallis and Rolls, 1997a) emerged in the late 1990s as a biologically plausible model that diverged from purely spatial accounts of vision by emphasizing the role of temporal continuity in stimulus sequences (Rolls, 2021a). The model captures how the brain processes consecutive visual inputs, interpreting them as different views of the same object under natural transformations such as scaling, rotation, or illumination change. Through this mechanism, VisNet learns transformation-invariant representations in a manner concep-tually related to Self-Organizing Maps (SOMs) (Kohonen, 1982), but with the added ability to learn from temporally sequential input patterns. VisNet integrates two core learning principles: the Hebbian rule (Hebb, 1949a), often summarized as "neurons that fire together, wire together," and the trace learning rule (Rolls and Stringer, 2006; Rolls, 2021a). The latter reinforces neural responses to stimuli that occur in close temporal proximity, increasing activation consistency when successive inputs likely represent the same object—an assumption biologically supported by natural visual experience. Combined, these mechanisms allow the network to form invariant object representations from dynamic sequences of views. This ability makes VisNet particularly suitable for recognizing symmetric objects since its temporal continuity mechanism naturally captures reflectional and rotational relationships among sequential stimuli. Subsequent studies (Rolls and Stringer, 2006; Rolls, 2021a) further validated VisNet's robustness for invariant object recognition. Rolls (Rolls, 2021a) investigated how the primate brain recognizes objects despite variations in position, lighting, and orientation by linking computational models to neurophysiological findings, particularly those involving the inferior temporal cortex (ITC)—a region critical for complex shape representation. The work demonstrated that hierarchical visual processing combined with learning from experience supports the abstraction of identity-preserving features. These insights reinforced VisNet's relevance for both computational neuroscience and artificial vision, where achieving invariance remains a central challenge.

## 3.1 Architecture and Learning Principles

VisNet is organized as a hierarchical four-layer network designed to emulate stages of cortical visual processing. Each layer corresponds to a distinct area of the visual pathway, where progressively larger receptive fields and increasing complexity of representation mirror biological organization. Figure 1 illustrates the VisNet architecture. The input layer encodes local visual features such as edges and contrast variations, analogous to neuronal responses in the primary visual cortex (V1). Subsequent layers gradually integrate these primitive features into more complex and stable object representations (Rolls, 2021a), enabling biologically plausible hierarchical learning and invariant recognition. Learning in VisNet is governed by two complementary principles:

- **Hebbian Learning:** The strength of synaptic connections increases proportionally to the correlation between presynaptic and postsynaptic activity. When a presynaptic neuron ($x_j$) and a postsynaptic neuron ($y$) fire simultaneously, the connection between them is reinforced, following the principle that "neurons that fire together, wire together."

- **Temporal Continuity:** Consecutive inputs occurring close in time are assumed to represent the same object under different transformations. This encourages temporal associations between views, supporting the learning of invariant representations for recognition.

The change in synaptic weight $\delta w_j$ for an input neuron $x_j$ is given by the trace learning rule (Rolls, 2021a):

$$\delta w_j = \alpha \bar{y}_\tau x_j, \tag{1}$$

where $\alpha$ is the learning rate and $\bar{y}_\tau$ is the temporal trace of the postsynaptic neuron's output at time step $\tau$, representing a history-weighted average of prior activations. The trace value updates according to:

$$\bar{y}_\tau = (1 - \eta) y_\tau + \eta \bar{y}_{\tau-1}, \tag{2}$$

where $\eta$ controls the relative weighting of the current response ($y_\tau$) versus previous outputs ($\bar{y}_{\tau-1}$). Higher values of $\eta$ emphasize prior activations, while lower values favor the most recent input. Rolls (Rolls, 2021a) reports optimal values of $\eta$ typically near 0.8, balancing memory persistence and adaptability. Together, these equations enable VisNet to associate temporally contiguous inputs, forming stable, invariant representations that facilitate recognition of objects across changes in orientation, scale, or position.

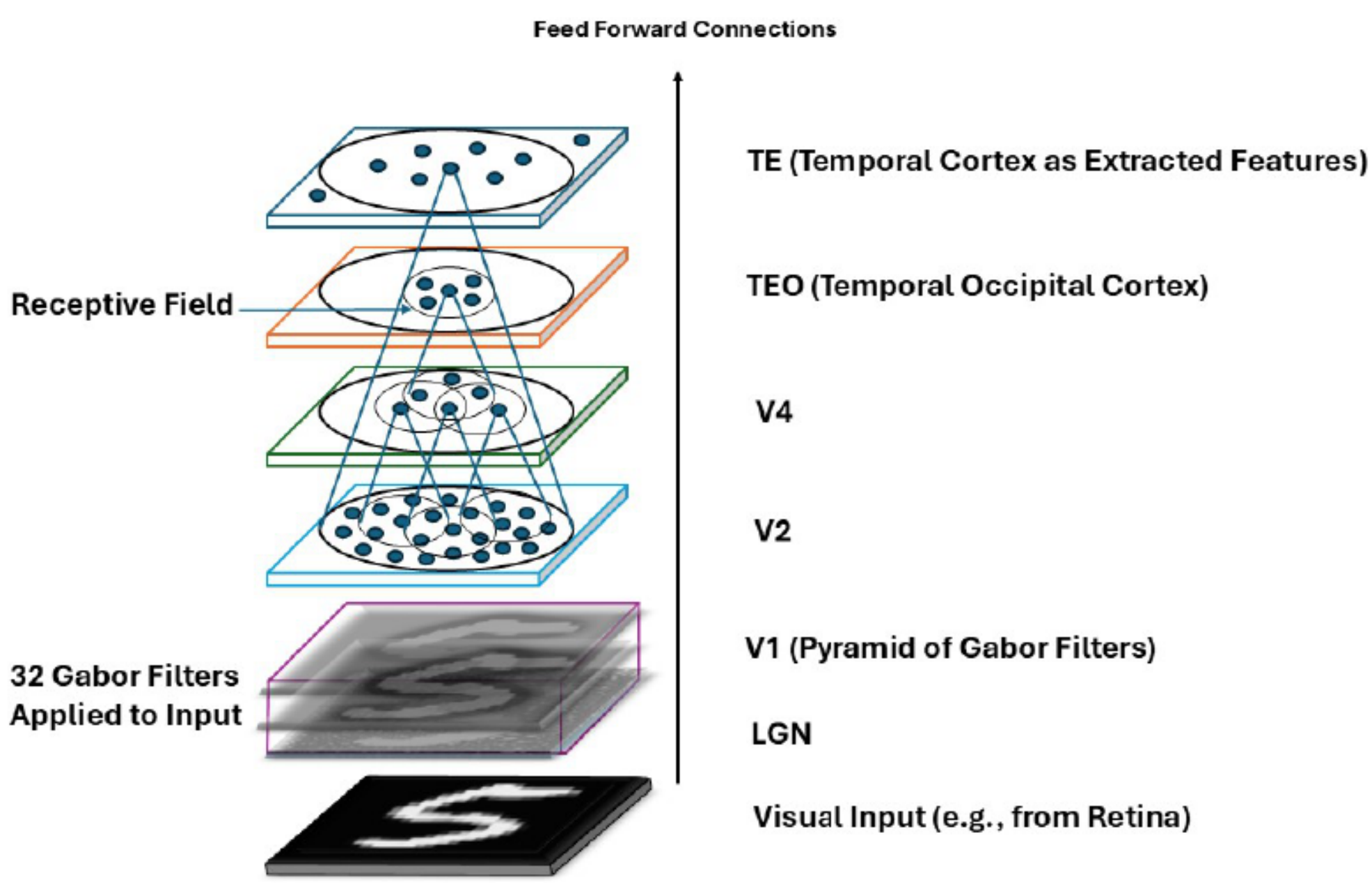

Figure 1: Schematic representation of the VisNet model, showing hierarchical layers and their correspondence to visual cortical areas (Rolls, 2021a).

## 3.2 Min–Max Normalization and Weight Stabilization

This study adopts the original parameterization proposed by Rolls (Rolls, 2021a), including the use of Gabor filters and neuron types across network layers. To prevent neuronal saturation—a common issue in Hebbian-based models—we employ a Min–Max normalization of neuron activations to maintain values within a bounded range $[0, 1]$. The normalized activation $y$ for input $x$ is computed as:

$$y = \frac{x - \min(x)}{\max(x) - \min(x)}, \tag{3}$$

where $\min(x)$ and $\max(x)$ represent dynamic bounds over the current input window. This normalization allows adaptive scaling of activations and ensures stable learning performance. In addition, synaptic weights are normalized after each update to preserve numerical stability and biological plausibility. Weight vectors are constrained using the following rule:

$$\mathbf{W}_{\text{normalized}} = \frac{\mathbf{W}_{\text{updated}}}{\|\mathbf{W}_{\text{updated}}\|}, \tag{4}$$

where $\|\mathbf{W}_{\text{updated}}\|$ denotes the vector norm of the updated weights. This ensures controlled magnitude of synaptic strengths and prevents divergence during training. By combining Min–Max normalization with weight stabilization, the model achieves robust convergence and consistency with neurobiological constraints.

# 4 Background: VisNet-Simplified and HMAX

## 4.1 VisNet-Simplified

VisNet-Simplified is a four-layer hierarchical neural model derived from the original VisNet architecture Rolls (2021a). It employs Hebbian learning combined with a temporal trace rule to develop invariant object representations from sequences of temporally contiguous inputs. This simplified version serves as the baseline configuration in our experiments. To reduce computational cost, especially given the intensive processing required by Gabor pyramid inputs at high resolutions (e.g., $256 \times 256$), the VisNet-Simplified model operates on $32 \times 32$ input images. The model omits sparsity constraints to maximize the utilization of small receptive fields, enabling efficient hierarchical processing. Through successive layers, it evolves from low-level edge detection in V1-like representations to higher-level, transformation-invariant object recognition in the final stages.

## 4.2 HMAX

The HMAX model is a hierarchical, feedforward architecture composed of alternating simple (S) and complex (C) layers that progressively build invariance to scale and translation Rolls (2015). Feature extraction in HMAX relies on multi-scale Gabor filters at the S-layers, followed by max-pooling operations at the C-layers to achieve position and scale tolerance. Unlike VisNet and its extensions, HMAX does not incorporate temporal learning or associative mechanisms, functioning purely as a static feedforward system. In this study, HMAX is included as a baseline model to benchmark the performance of our proposed biologically inspired architectures.

# 5 Enhanced VisNet-Simplified Variants

## 5.1 Incorporating RBF Neurons into VisNet-Simplified (VisNet-RBF)

Incorporating Radial Basis Function (RBF) neurons into VisNet-Simplified offers a biologically plausible alternative to traditional fully connected McCulloch-Pitts neurons for certain tasks. RBF neurons rely on a Gaussian activation function, where the output decreases as the input moves away from a center or prototype vector. This mechanism mimics localized response characteristics, which can be beneficial for recognizing patterns or objects, especially when working with symmetric structures, such as those explored in Bishop (1995). The localized nature of RBF allows for more precise feature detection in specific regions of the input space, which is particularly advantageous in symmetry tasks where local alignments are critical.

### 5.1.1 RBF Neurons and Gaussian Activation

The most common RBF activation function used is the Gaussian function Bishop (1995), which is expressed as:

$$\phi(\mathbf{x}) = \exp\left(-\frac{||\mathbf{x} - \mathbf{c}||^2}{2\sigma^2}\right) \tag{5}$$

Where:

- $\mathbf{x}$ is the input vector,
- $\mathbf{c}$ is the center vector (prototype),
- $\sigma$ controls the width of the receptive field.

This function results in a localized response that is strongest when the input $\mathbf{x}$ is close to the center $\mathbf{c}$, which represents the weight vector for each neuron in VisNet-Simplified. As the input moves further from the center, the output of the neuron decreases, enabling the network to be sensitive to specific patterns. In the context of VisNet-Simplified, this feature is useful for learning and recognizing objects, as it emphasizes local features that are critical for detecting patterns.

### 5.1.2 Motivation for Incorporating RBF Neurons

The inclusion of RBF neurons into VisNet-Simplified is motivated by their ability to capture localized features efficiently. Akbarinia et al. Akbarinia et al. (2017); Parraga et al. (2019) demonstrated the effectiveness of low-level operators for symmetry detection using Gabor filters to extract symmetry axes from simple figures. These operators, like RBF neurons, emphasize local symmetry features, offering a computationally efficient mechanism for pattern recognition. Integrating RBF neurons with VisNet-Simplified extends this principle by incorporating a Gaussian activation function, which is biologically plausible and computationally robust for tasks involving symmetry. By drawing on these principles, VisNet-RBF is positioned as an enhanced model for symmetry detection, leveraging localized responses to better handle symmetric and complex visual patterns.

## 5.2 Improved VisNet-Simplified Model with Mahalanobis Distance (VisNet-MD)

The original VisNet-Simplified model, while powerful in handling complex visual patterns, suffers from a saturation problem where the network struggles to generalize effectively in high-dimensional or noisy data scenarios (Olshausen and Field, 1996). This limitation can hinder the ability of VisNet-Simplified to maintain stable, invariant representations, particularly when faced with data that is sparse or imbalanced. Additionally, the Hebbian learning rule, while simple and biologically inspired, has limitations in terms of its accuracy and scalability, as it does not take into account the complex relationships between features in high-dimensional data. Hebbian learning strengthens the connections between co-activated neurons, but it does not provide a mechanism for adjusting to variations in data or improving learning accuracy in more complex tasks Hebb (1949a). To address these weaknesses, the VisNet-Simplified model can be enhanced by integrating an unsupervised learning mechanism that utilizes the gradient of the Mahalanobis distance Mahalanobis (1936). This allows the network to learn representations based on the statistical properties of the input data, improving its ability to adapt to various visual transformations. The Mahalanobis distance is particularly well-suited for improving VisNet-Simplified in unsupervised learning due to its ability to account for correlations between features by using the covariance matrix, making it more robust than traditional Euclidean distance Fukunaga (1990). Unlike Euclidean distance, Mahalanobis distance is scale-invariant, ensuring consistent learning even when input features have varying magnitudes. Additionally, it is less sensitive to outliers, which enables the model to focus on meaningful patterns and improves its ability to recognize consistent features despite noise Olshausen and Field (1996). Furthermore, Mahalanobis distance adapts well to elliptical clusters, which is reflective of the natural distribution of real-world data. This characteristic enhances discriminability by emphasizing covariance differences between object categories, improving the ability of the model to distinguish between similar objects Kaiming He (2016). This adaptability to high-dimensional data ensures more effective learning of invariant representations in VisNet-Simplified, especially in complex visual processing tasks Yann LeCun (1998). The combination of these features helps overcome the limitations imposed by saturation, improving the performance and generalization of VisNet-Simplified in both supervised and unsupervised learning scenarios.

### 5.2.1 Mahalanobis Distance

Mahalanobis Distance (MD) is a multivariate measure of distance that accounts for correlations between variables. It is especially useful when data features are correlated or have unequal variances. Unlike Euclidean distance, which computes the straight-line distance between two points, Mahalanobis Distance measures the distance between a point and a distribution, considering the distribution's covariance structureMahalanobis (1936). The Mahalanobis distance between a data point and a mean vector with covariance matrix is defined as:

$$D_M(\mathbf{x}, \boldsymbol{\mu}) = \sqrt{(\mathbf{x} - \boldsymbol{\mu})^\top \boldsymbol{\Sigma}^{-1} (\mathbf{x} - \boldsymbol{\mu})} \tag{6}$$

where:

- $\mathbf{x}$ is the vector representing the data point.

- $\mu$ is the mean of the distribution.
- $\Sigma^{-1}$ is the inverse of the covariance matrix of the distribution.
- $(\mathbf{x} - \mu)$ is the difference between the data point and the mean, indicating the deviation.

This distance metric takes into account the correlations of the data set and scales the distances accordingly.

### 5.2.2 Gradient Learning

To facilitate unsupervised learning, we consider the gradient of the Mahalanobis distance with respect to the weights of the synaptic connections in the network. The update rule for the synaptic weight can be expressed as follows:

$$\delta w_j = -\alpha \nabla D_M(\mathbf{x}, \boldsymbol{\mu}) \tag{7}$$

where $\alpha$ is the learning rate and $\nabla D_M(\mathbf{x}, \boldsymbol{\mu})$ is the gradient of the Mahalanobis distance.

### 5.2.3 Gradient Calculation

The gradient of the Mahalanobis distance with respect to the weights can be computed as:

$$\nabla D_M(\mathbf{x}, \boldsymbol{\mu}) = \frac{1}{D_M(\mathbf{x}, \boldsymbol{\mu})} \left( \boldsymbol{\Sigma}^{-1} (\mathbf{x} - \boldsymbol{\mu}) \right) \tag{8}$$

This gradient informs the model how to adjust the weights to minimize the Mahalanobis distance, effectively improving the learning capability of the network in an unsupervised manner.

### 5.2.4 Overall Learning Rule

Combining the original synaptic weight update with the Mahalanobis distance learning, we obtain the updated weight rule as follows:

$$\delta w_j = \alpha (\nabla D_M(\mathbf{x}, \boldsymbol{\mu}) - w_j) \tag{9}$$

Here, represents the output from the neuron at time step , enabling the network to learn from both the output activations and the statistical relationships captured by the Mahalanobis distance.

# 6 Imitation from Local Inhibition in the Visual Cortex (VisNet-LI)

The visual cortex is a cornerstone of our understanding of biological vision, not only due to its laminar structure but also because of its columnar organization. Columns, such as orientation and ocular dominance columns, are vertically aligned structures that traverse the cortical layers, systematically organizing neurons with shared functional properties. These properties include sensitivity to specific orientations, spatial frequencies, and eye-specific inputs Hubel and Wiesel (1962b). This columnar arrangement ensures efficient and structured encoding of visual stimuli, reflecting an intricate biological architecture optimized for processing diverse visual inputs. At the core of this functionality lies Hebbian learning, a synaptic plasticity mechanism that encapsulates the idea that "neurons that fire together, wire together" Hebb (1949a). Within the columnar framework, Hebbian learning enhances synaptic connections between neurons that consistently exhibit correlated activity. This not only facilitates the development of specialized neural responses but also supports the formation of hierarchical representations across successive cortical layers Rolls (2021a); Kohonen (1982). By integrating the spatial and functional relationships within columns, this learning principle underpins key features of visual processing, including edge detection, contour integration, and orientation tuning. Recent research underscores the efficiency of columnar organization in feature extraction and object recognition. Columns enable the hierarchical processing of spatially and temporally correlated features, ensuring robust representation of complex objects under varying transformations Riesenhuber and Poggio (1999); Rolls (2021a). For instance, orientation columns aid in encoding edges at different angles, which are further integrated to form higher-level shapes and patterns. Inspired by these biological principles, our proposed VisNet models incorporate a columnar-inspired structure to enhance their learning capabilities Mountcastle (1997). By adapting Hebbian learning to operate within a cylindrical organization spanning multiple layers, we enable the model to capture both hierarchical and spatial relationships in visual data Hebb (1949a). This approach mirrors the biological integration observed in the visual cortex, where receptive fields within columns influence neurons across layers Hubel and Wiesel (1962a). The resulting framework facilitates more robust learning of invariant object representations, enhancing the model's biological plausibility and effectiveness in dynamic visual tasks William R. Lindsay (2010).

# 7 VisNet-Li-DoG-RGB-WTA

The **VisNet-Li-DoG-RGB-WTA** model is an enhanced, biologically inspired extension of the VisNet-Li architecture Rolls et al. (1998); Wallis and Rolls (2001). It introduces a dual-stage preprocessing pipeline combining *Difference of Gaussian (DoG)* filtering and multi-scale Gabor pyramids to emulate computations of the retina and primary visual cortex (V1). This approach enriches early visual representations with luminance, chromatic, and orientation-selective information, further refined by a discrete **Winner-Take-All (WTA)** selection mechanism.

## 7.1 DoG-Based Retinal Preprocessing

RGB images are first transformed into opponent channels to mimic biological color processing:

$$L(x,y) = \frac{R(x,y) + G(x,y) + B(x,y)}{3}, \quad R_G(x,y) = R(x,y) - G(x,y), \quad B_G(x,y) = B(x,y) - G(x,y), \tag{1}$$

where $L$ represents luminance, and $R_G, B_G$ represent chromatic contrasts. Each channel is filtered using a Difference-of-Gaussian kernel modeling retinal ganglion center–surround receptive fields Marr and Hildreth (1980):

$$\mathrm{DoG}(x,y) = G(x,y,\sigma_1) - k \cdot G(x,y,\sigma_2), \quad G(x,y,\sigma) = \frac{1}{2\pi\sigma^2} \exp\left(-\frac{x^2+y^2}{2\sigma^2}\right), \tag{2}$$

with parameters $\sigma_1 = 1.0$, $\sigma_2 = 1.2$, and $k = 0.6$. The resulting output is a three-channel tensor $I_{\mathrm{DoG}}(x,y) = [L_{\mathrm{DoG}}, R_G^{\mathrm{DoG}}, B_G^{\mathrm{DoG}}]$.

## 7.2 Channel-Wise Gabor Pyramid Construction

Each DoG channel is processed through multi-scale, multi-orientation Gabor filters to simulate V1 simple cells:

$$G_{\mathrm{Gabor}}(x,y) = \exp\left(-\frac{x'^2 + \gamma^2 y'^2}{2\sigma^2}\right) \cos(2\pi f x' + \phi), \tag{3}$$

where $x'$ and $y'$ are the rotated coordinates. The independent pyramids are concatenated: $I_{\mathrm{combined}} = [P_L, P_{R_G}, P_{B_G}]$, forming a unified feature tensor for hierarchical processing.

## 7.3 Early Visual Processing: DoG and Gabor Filtering

Prior to hierarchical processing, the input images undergo biologically-inspired preprocessing that mimics retinal and V1 cortical filtering operations. This preprocessing extracts fundamental visual primitives essential for downstream feature learning.

### 7.3.1 Difference of Gaussians (DoG) Filtering

Following retinal center-surround receptive field organization Enroth-Cugell and Robson (1966), we apply a Difference of Gaussians (DoG) filter to extract luminance and color-opponent channels:

$$\mathrm{DoG}(x,y) = \frac{1}{2\pi\sigma_1^2} e^{-\frac{x^2+y^2}{2\sigma_1^2}} - 0.6 \cdot \frac{1}{2\pi\sigma_2^2} e^{-\frac{x^2+y^2}{2\sigma_2^2}} \tag{4}$$

where $\sigma_1 = 1.0$ (center) and $\sigma_2 = 1.2$ (surround) define the spatial scales, and the coefficient 0.6 balances the antagonistic surround contribution. The filter is applied to three distinct channels:

$$\mathbf{L} = \mathrm{DoG} * \frac{R+G+B}{3} \qquad \text{(Luminance)} \tag{5}$$

$$\mathbf{C}_1 = \mathrm{DoG} * (R - G) \qquad \text{(Red-Green opponency)} \tag{6}$$

$$\mathbf{C}_2 = \mathrm{DoG} * (B - G) \qquad \text{(Blue-Green opponency)} \tag{7}$$

where $*$ denotes 2D convolution. Each channel undergoes per-image min-max normalization to ensure numerical stability:

$$\mathbf{L}_{\text{norm}} = \frac{\mathbf{L} - \min(\mathbf{L})}{\max(\mathbf{L}) - \min(\mathbf{L}) + \epsilon}, \quad \epsilon = 10^{-8} \tag{8}$$

#### 7.3.2 Gabor Filter Bank

To capture oriented edge and texture information analogous to V1 simple cells Hubel and Wiesel (1962a), we employ a bank of Gabor filters with systematic parameter variation:

$$G(x, y; f, \theta, \phi) = \frac{1}{\sqrt{2\pi}} \cdot 2^{-(f-0.25)} \cdot e^{-\frac{4x'^2+y'^2}{8}} \cdot \cos(\pi x' + \phi) \tag{9}$$

where rotated coordinates are defined as:

$$\begin{bmatrix} x' \\ y' \end{bmatrix} = \begin{bmatrix} \cos\theta & \sin\theta \\ -\sin\theta & \cos\theta \end{bmatrix} \begin{bmatrix} 2^{-(f-0.25)}x - s/2 \\ 2^{-(f-0.25)}y - s/2 \end{bmatrix} \tag{10}$$

with $s$ being the spatial filter size. The filter bank spans:

- **Frequencies**: $f \in \{0.25, 0.5, 1.0, 2.0\}$ cycles per image (filter sizes: $7 \times 7$, $11 \times 11$, $15 \times 15$, $19 \times 19$)
- **Orientations**: $\theta \in \{0, 45, 90, 135\}$
- **Phases**: $\phi \in \{0, \pi/2\}$ (even and odd symmetric)

This yields $4 \times 4 \times 2 = 32$ complex Gabor filters. The complex response is computed as:

$$R_{\text{Gabor}} = \sqrt{(G_{\text{real}} * \mathbf{C})^2 + (G_{\text{imag}} * \mathbf{C})^2} \tag{11}$$

where $\mathbf{C} \in \{\mathbf{L}, \mathbf{C}_1, \mathbf{C}_2\}$. Applying all 32 Gabor filters to each of the 3 DoG channels produces $32 \times 3 = 96$ feature channels that serve as input to the hierarchical network.

#### 7.3.3 Adaptive Lateral Plasticity via Hebbian Learning on Inhibitory Connections

In hierarchical models of visual processing, such as variants of the VisNet architecture, lateral inhibition within layers is essential for enforcing competition, sparsifying representations, and promoting feature selectivity. While traditional VisNet implementations rely on fixed lateral inhibition to achieve decorrelation and winner-take-all dynamics, here we introduce an adaptive lateral plasticity mechanism that dynamically modulates inhibitory connections based on neuronal co-activation patterns. This Hebbian-based rule on inhibitory synapses enables selective disinhibition among frequently co-active neurons, leading to the emergent formation of cooperative ensembles. Empirical results on VisNet variants demonstrate that this adaptive inhibition significantly improves classification accuracy on benchmark visual recognition tasks compared to fixed-inhibition baselines, while preserving biological plausibility and enhancing representational modularity. The lateral connectivity matrix $\mathbf{W}^{\text{lat}} \in \mathbb{R}^{N_\ell \times N_\ell}$ evolves via a *Hebbian plasticity rule applied to inhibitory synapses*. Crucially, while the weights themselves are inhibitory (negative-valued), the learning rule follows standard Hebbian dynamics where co-activation *reduces* mutual inhibition:

$$\Delta\mathbf{W}^{\text{lat}} = \frac{\eta}{B}\left(\mathbf{y}^\top\mathbf{y} - \delta_{\text{inh}}\,\mathbf{W}^{\text{lat}}\right) \tag{12}$$

where $\mathbf{y} \in \mathbb{R}^{B \times N_\ell}$ represents batch activities, and $\mathbf{y}^\top \mathbf{y}$ captures pairwise correlations. Since $\mathbf{W}^{\text{lat}}$ is initialized with negative values ($W_{ij}^{\text{lat}}(0) = -0.1$), the effect is:

- **Co-active neuron pairs**: $\mathbf{y}_i^\top \mathbf{y}_j > 0 \Rightarrow \Delta W_{ij}^{\text{lat}} > 0 \Rightarrow$ weight becomes less negative $\Rightarrow$ **reduced inhibition** (cooperative ensemble formation)
- **Anti-correlated pairs**: $\mathbf{y}_i^\top \mathbf{y}_j \approx 0 \Rightarrow$ weight remains strongly negative $\Rightarrow$ **maintained inhibition** (competitive decorrelation)

This implements a form of **Hebbian disinhibition** Letzkus et al. (2015); Pi et al. (2013), where the learning rule itself is excitatory (cells that fire together wire together) but operates on inhibitory connections, leading to the emergence of cooperative neural ensembles that mutually reduce their reciprocal inhibition. The mechanism differs from classical anti-Hebbian learning (where co-activation would *strengthen* inhibition) and instead implements a biologically-observed phenomenon where synchronized activity leads to reduced mutual suppression, facilitating the formation of functional cell assemblies Buzsáki (2010). This adaptive lateral inhibition mechanism, based on Hebbian plasticity applied to inhibitory synapses, offers a biologically plausible and functionally advantageous approach to sculpting recurrent dynamics in neural networks. By selectively reducing mutual inhibition among co-active neurons—effectively implementing Hebbian disinhibition Letzkus et al. (2015); Pi et al. (2013)—the rule enables the emergent formation of cooperative cell assemblies that can sustain coordinated activity, as observed in cortical circuits where synchronized firing leads to reduced suppression and enhanced ensemble persistence Buzsáki (2010). Unlike fixed lateral inhibition, which imposes uniform competition without adaptability, or classical anti-Hebbian rules on inhibitory connections that would strengthen suppression for co-active pairs (enforcing stricter decorrelation at the cost of assembly formation), this method dynamically balances cooperation within ensembles and competition across them. The result is improved representational capacity, with modular, sparse activity patterns that support robust feature binding and self-organization, making it particularly suitable for unsupervised learning in large-scale recurrent models inspired by neocortical processing.

# 8 Methodology

## 8.1 Dataset

In this study, several datasets were employed to evaluate VisNet's object recognition capabilities and to establish a foundation for future research on symmetry ranking. Each dataset was selected to assess different aspects of the model's performance, ranging from simple grayscale classification tasks to complex multi-class and symmetry-based recognition challenges. A summary of all datasets, including their sizes, resolutions, and intended roles within the experiments, is presented in Table 1.

| Dataset | Number of Images | Image Size | Description |
|---|---|---|---|
| **Caltech-256** | 30,607 | $32 \times 32$ (resized from $256 \times 256$) | A diverse collection of images across 256 object categories, resized from the original resolution of $256 \times 256$ to $32 \times 32$ pixels due to the computational demands of the Gabor pyramid. This dataset was used to examine VisNet-Simplified's performance on real-world objects with varied appearances and orientations. |
| **MNIST** | 70,000 | $28 \times 28$ | A benchmark dataset of grayscale handwritten digits used to evaluate VisNet-Simplified's ability to recognize simple, uniform shapes under controlled conditions. |
| **CIFAR-10** | 60,000 | $32 \times 32$ | A dataset of RGB images distributed across ten object categories, providing a challenging testbed for evaluating VisNet-Simplified's performance on colored, natural scenes containing diverse objects. |
| **Custom Symmetric Sets** | Variable | Varies (binary and RGB) | A custom-designed collection of binary and RGB images—including squares, Sierpiński triangles, and human-like figures—exhibiting varying levels of symmetry. This dataset was developed to investigate VisNet's capacity for symmetry detection and to support future work on symmetry ranking. |

Table 1: Overview of the datasets used in this study, including their image counts, resolutions, and specific roles in evaluating VisNet-Simplified's performance in object recognition and symmetry analysis.

#### 8.1.1 Dataset for Symmetry Recognition in Degraded Square Shapes

To evaluate the model's ability to recognize approximate symmetry, we constructed a dataset of binary images depicting square-shaped objects with varying levels of degradation. Each level introduced controlled asymmetry, allowing an examination of the model's sensitivity to progressive structural distortions (Figure 2). This setup simulated different degrees of real-world symmetry degradation, challenging the model to extract invariant geometric features despite partial occlusion or noise. For simplified evaluation, a two-class variant (TWOCLASSES-SQUARE) was employed, containing only the first and fifth symmetry levels.

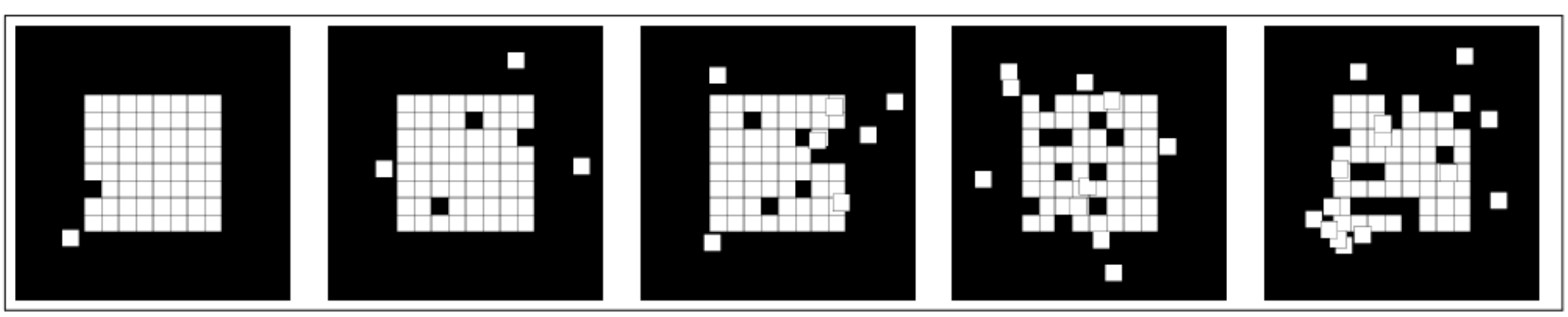

Figure 2: Binary square images representing five symmetry levels (SQUARE dataset).

### 8.1.2 Sierpiński Triangle and Symmetric Object Generation

The Sierpiński triangle, a recursive geometric fractal composed of equilateral triangles, was used to generate symmetric objects for further experimentation. Its self-similar properties at successive levels of recursion make it an ideal candidate for exploring symmetry perception in computational models. Figure 3 illustrates sample objects from this dataset, which were used to assess the model's ability to interpret hierarchical and fractal symmetry patterns.

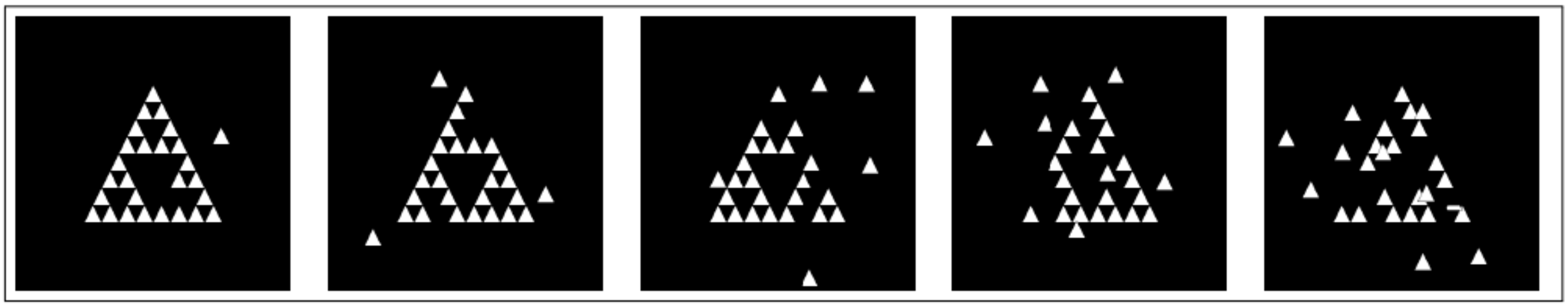

Figure 3: Example objects from the Sierpiński Triangle dataset, depicting five symmetry levels (TRIANGLE).

### 8.1.3 Robustness Testing with Rotated and Translated Triangles

To evaluate rotational and positional invariance, additional experiments were performed using rotated and translated Sierpiński triangles. Rotations were applied within a range of $[-180°, 180°]$, while translations were constrained to $[-20\%, 20\%]$ of the image dimensions. The model successfully recognized objects across these transformations, demonstrating robust symmetry detection under varied viewing conditions (ROTATED-TRANSLATED-TRIANGLE).

### 8.1.4 Symmetry Recognition in Detached and Reattached Squares

To test VisNet's sensitivity to rule-based symmetry, a dataset of binary square objects was created in which object segments were deliberately detached and reattached following predefined symmetry principles. This design emphasized symmetry as the primary discriminative feature and challenged the model to classify objects using geometric regularity rather than local shape. Example stimuli are shown in Figure 4, and classification outcomes are reported in Table 3. A simplified two-class version, denoted as TWOCLASSES-PARTED-SQUARE, was also employed to evaluate the model's ability to separate symmetric and asymmetric categories.

### 8.1.5 Evaluating Symmetry Recognition with Split-Categorized Square Objects

An extended version of the binary square dataset was generated by introducing categories based on the number of splits, producing objects with more nuanced symmetry variations. This variant enabled a deeper examination of the model's capacity to generalize across differing symmetry complexities. Figure 5 shows representative examples from this dataset. Performance results for both two-class (TWOCLASSES-SOMEPARTED-SQUARE) and five-class (FIVECLASSES-SOMEPARTED-SQUARE) configurations are reported in Table 3.

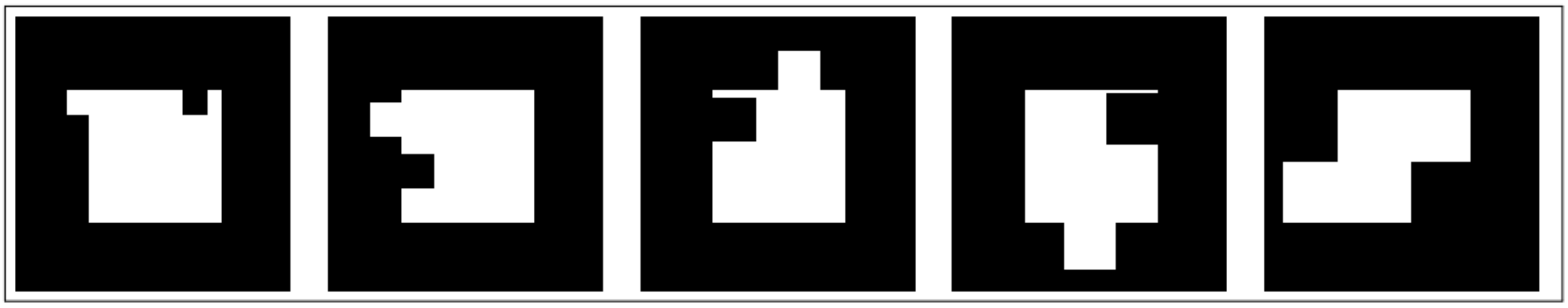

Figure 4: Binary square images with five symmetry levels (FIVECLASSES-PARTED-SQUARE).

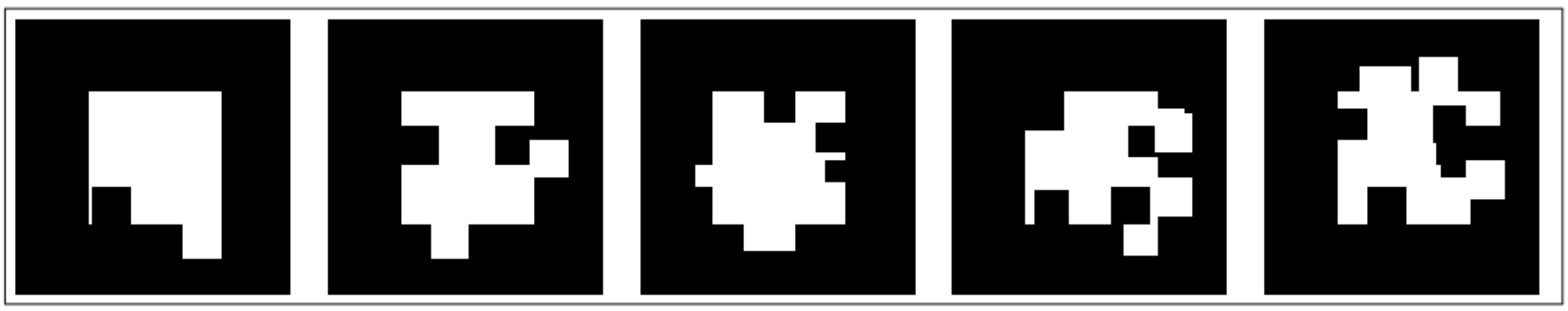

Figure 5: Representative images from the FIVECLASSES-SOMEPARTED-SQUARE dataset.

#### 8.1.6 Challenging Models with Symmetry in Complex Human-Like Objects

Building on the previous datasets, a more intricate collection of binary, human-like shapes was designed, each characterized by five distinct symmetry levels. As in earlier datasets, portions of each object were detached and reattached according to specified symmetry rules. Rotations and translations were also introduced to increase variability, producing a challenging test for visual invariance. Figure 6 presents example images, and corresponding classification results are summarized in Table 3. This dataset was specifically developed to train and evaluate the model's ability to recognize symmetry as the defining feature, independent of other structural details.

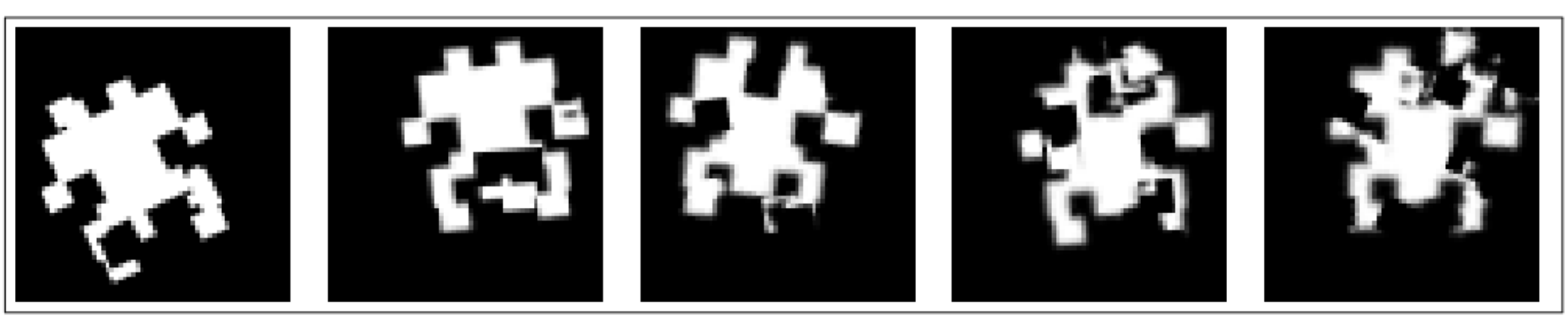

Figure 6: Binary human-like objects exhibiting five symmetry levels (ROTATED-TRANSLATED-HUMAN-LIKE).

### 8.1.7 RGB Dataset of Symmetric Objects with Varying Levels of Symmetry

An additional RGB dataset was developed to explore VisNet's color sensitivity in symmetry perception. It contains objects with five distinct symmetry levels rendered in varied backgrounds. Each class corresponds to a specific degree of symmetry, ensuring internal consistency within class samples. Figure 7 displays representative examples, arranged from low (20%) to high (100%) symmetry.

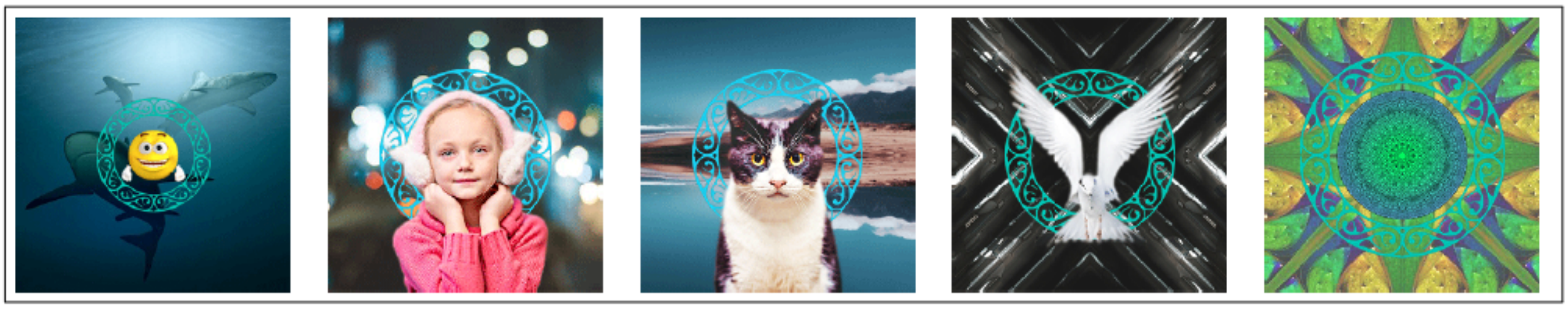

Figure 7: RGB images depicting five symmetry levels (RGB-IMAGE dataset). From left to right, symmetry increases from 20% to 100%.

### 8.1.8 MNIST: A Benchmark Dataset for Machine Learning and Computer Vision

The MNIST (Modified National Institute of Standards and Technology) dataset is a standard benchmark for evaluating image recognition models. It comprises 70,000 grayscale images of handwritten digits (0–9), each of size $28 \times 28$ pixels, divided into 60,000 training and 10,000 test images. MNIST provides a controlled environment for assessing VisNet's performance in recognizing simple, well-defined patterns. Example digits are shown in Figure 8.

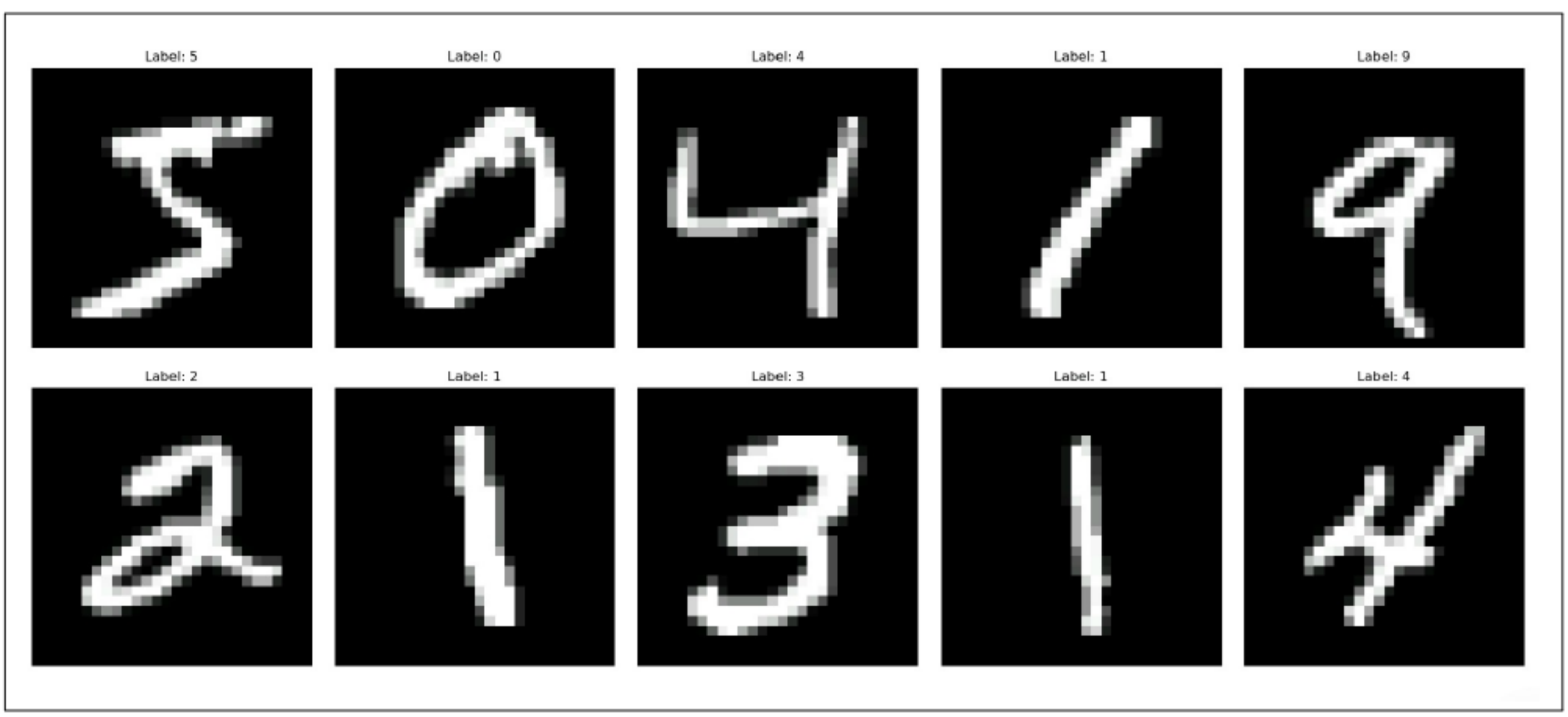

Figure 8: Sample MNIST digits used for benchmarking Lecun et al. (1998).

### 8.1.9 CIFAR-10: A Comprehensive Benchmark for Visual Object Categories

The CIFAR-10 dataset is widely used for evaluating image classification models. It contains 60,000 RGB images of size $32 \times 32$ pixels, categorized into ten classes (*airplane, automobile, bird, cat, deer, dog, frog, horse, ship, truck*). The dataset is divided into 50,000 training and 10,000 test images, evenly distributed across classes. Its compact yet visually diverse structure makes it suitable for testing models that aim to balance computational efficiency with recognition capability. Representative samples are shown in Figure 9.

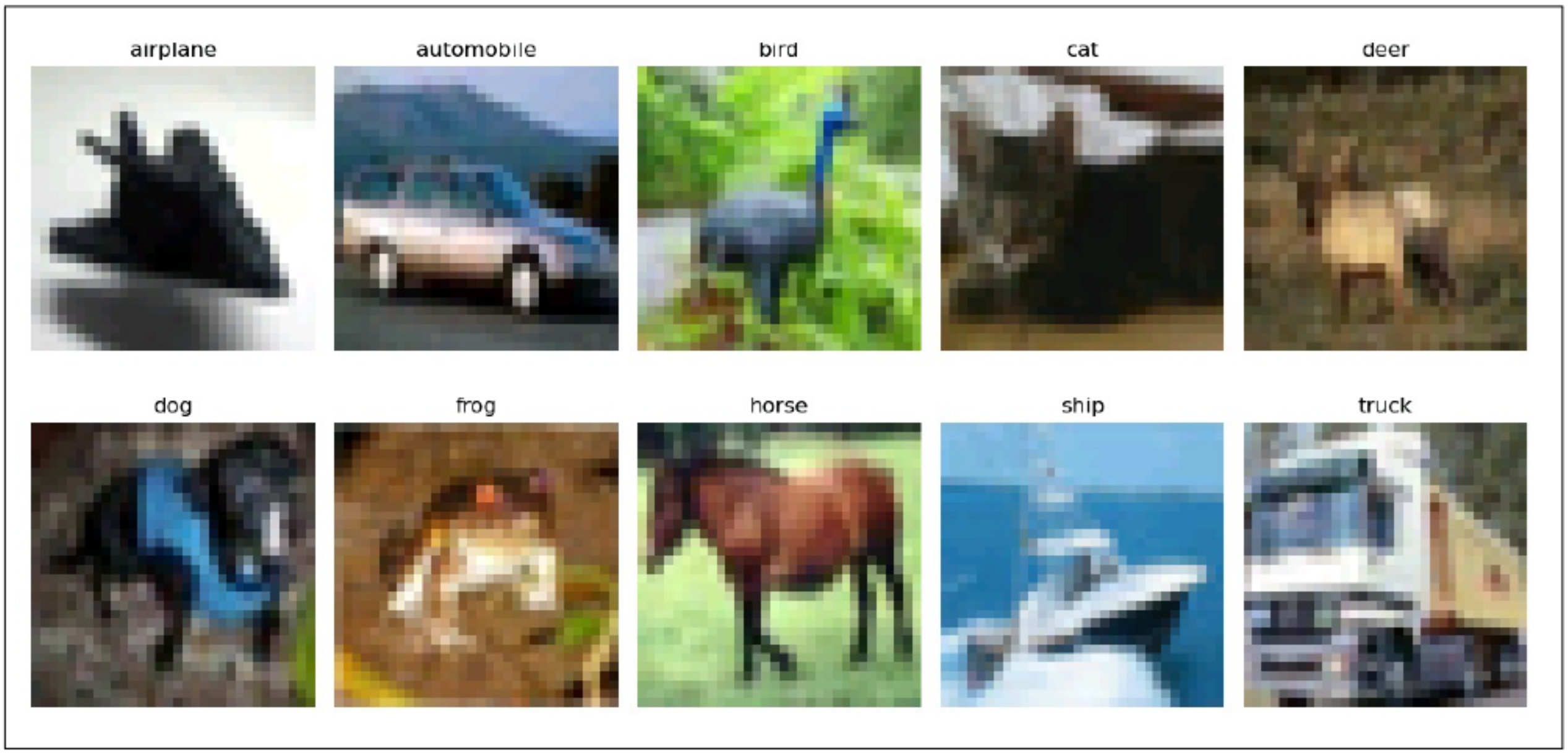

Figure 9: Example CIFAR-10 images across ten object categories Krizhevsky (2009).

## 8.2 Training

We trained the models using several datasets, each designed to evaluate different aspects of object recognition and symmetry perception. The input images were first processed by the initial layer, which extracted basic edge and contrast information. As the signals propagated through subsequent layers, the learning mechanism iteratively adjusted the synaptic weights, enabling the formation of progressively invariant representations and supporting robust object recognition. In the unsupervised training paradigm, random selection of samples during each iteration played a critical role. This stochastic exposure to diverse data points encouraged the model to learn meaningful feature patterns without relying on labeled examples. Random sampling also reduced the likelihood of overfitting by preventing the model from developing bias toward specific instances, thus promoting better generalization to unseen data. Each dataset contained 10,000 images, split into 80% for training and 20% for testing. A separate validation set was not required because, in competitive architectures such as VisNet, overfitting does not occur in the same way as in supervised models. Since there is no explicit hyperplane optimization and neurons compete through self-organization, the network inherently mitigates overfitting during training. This direct train–test division therefore provides an unbiased estimate of the model's generalization ability. Training was performed over 10 independent iterations, and the reported results represent the averages across these runs to ensure robustness and consistency. Prior to training, all RGB inputs were converted to grayscale and normalized to standardize the data distribution. After training, the synaptic weights within the receptive fields offered valuable insight into how the network encodes visual information across its hierarchy. Figure 10 illustrates representative receptive field patterns, showing the model's ability to extract and integrate visual features—from simple edges and textures in the lower layers to more complex and abstract patterns in the higher layers. In the computational model, a Gabor filter bank emulates the processing functions of the early visual system—particularly the lateral geniculate nucleus and primary visual cortex (LGN+V1) as described in Rolls (2021a). Each Gabor filter, tuned to different orientations, spatial frequencies, and positions, is designed to respond selectively to distinct texture and edge features. Through this mechanism, complex visual patterns are decomposed into simpler, localized components, mirroring the hierarchical feature extraction observed in biological vision. Figure 11 illustrates the conceptual analogy between the Gabor filter bank and LGN+V1 processing, emphasizing the evolutionary advantage of neurons sensitive to multiple orientations and spatial scales. This biologically inspired design demonstrates how sophisticated visual representations can emerge from the combination of numerous simple, specialized filters. The Gabor filter bank employed in this model is parameterized to capture a broad spectrum of visual characteristics. Spatial frequencies of $[0.25, 0.5, 1.0, 2.0]$ enable detection of textures across progressively finer scales. Orientations of $[0, \frac{\pi}{4}, \frac{\pi}{2}, \frac{3\pi}{4}]$ allow the filters to detect edges in horizontal, diagonal, vertical, and anti-diagonal directions. Phases of $[0, \pi]$ introduce phase shifts in the sinusoidal component, enabling sensitivity to patterns of differing alignments. Collectively, this combination of frequency, orientation, and phase parameters allows the Gabor bank to extract diverse texture and edge features from input stimuli, effectively simulating the spatial and angular tuning properties of the early visual system.

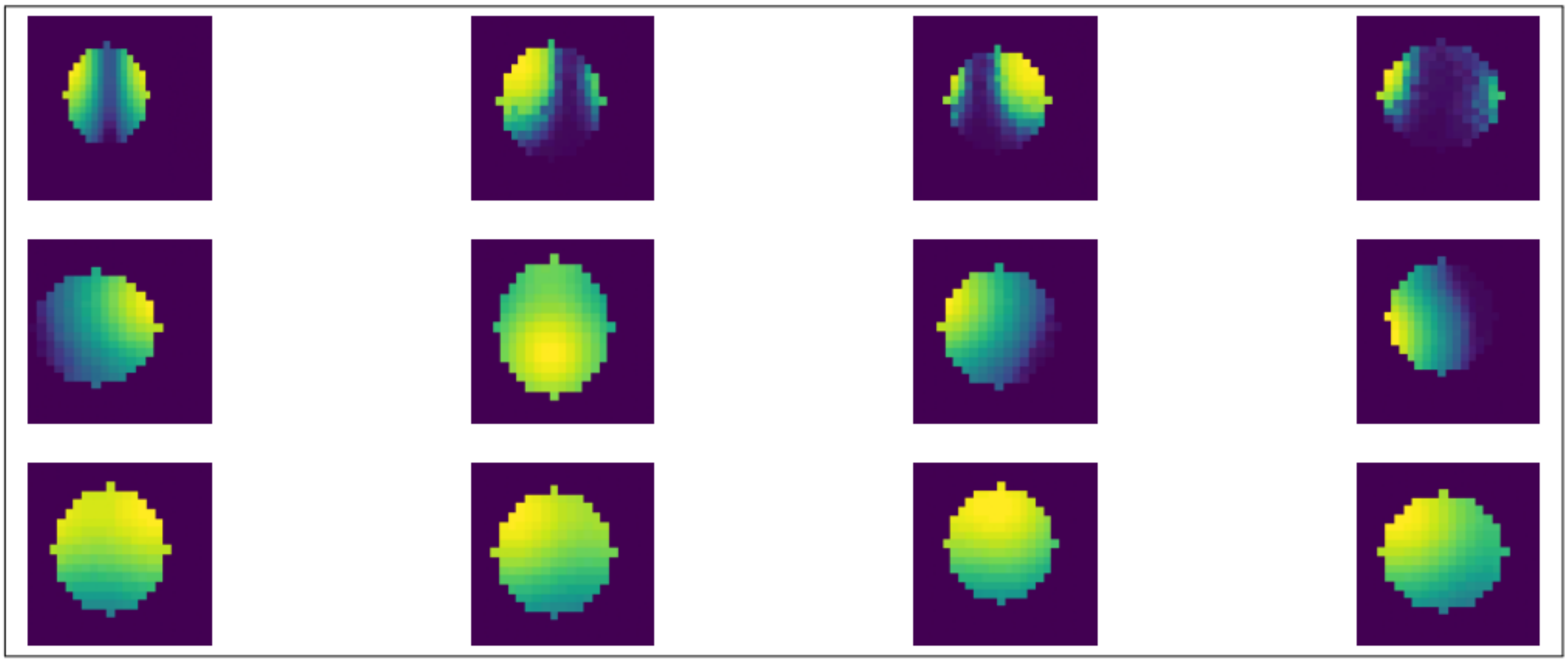

Figure 10: Visualization of some receptive fields after training. First row of images belong to first layer and second and third rows for second and third layer accordingly.

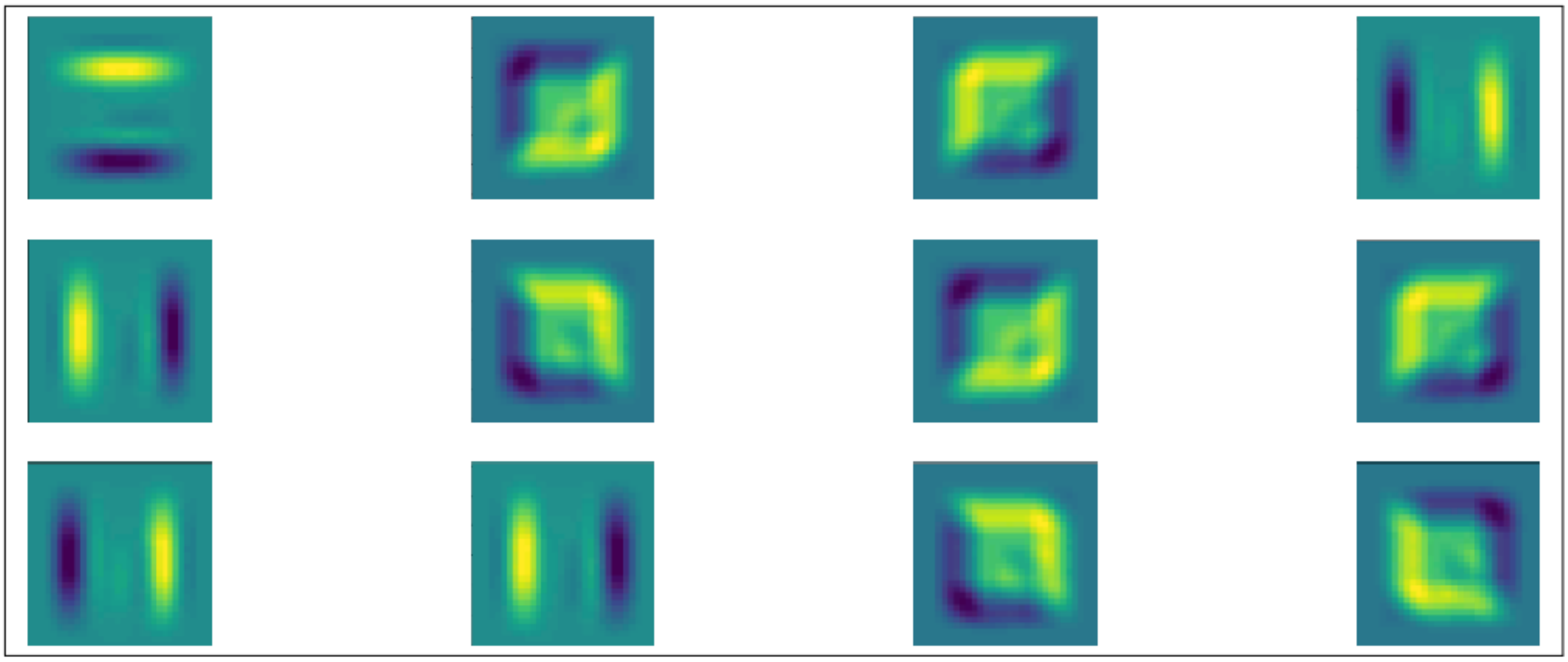

Figure 11: Results of applying Gabor filters on a square binary stimulus.

### 8.3 Testing and Classification

After training, we evaluated VisNet's capacity to recognize previously unseen symmetric objects by measuring its classification accuracy on novel test samples. The network's robustness was further assessed by applying a range of visual transformations, including scaling and rotation, to examine its ability to maintain performance under altered viewing conditions. In addition, standard benchmark datasets such as CIFAR-10 and MNIST were employed to test VisNet's generalization across diverse image types and varying levels of visual complexity. The following section presents and discusses the experimental findings, which are summarized in Table 3.

## 9 Results

The enhanced VisNet variants demonstrate robust performance across both standard benchmarks and controlled symmetry-based tasks. On MNIST, VisNet-MD achieves an accuracy of **94%**, while VisNet-LI-DoG-RGB-WTA reaches **52%** on CIFAR-10, substantially outperforming the baseline VisNet-Simplified (25%) and the classical HMAX model under comparable experimental conditions (Rolls, 2015). On several structured symmetry datasets (e.g., RGB-IMAGE and TWOCLASSES-PARTED-SQUARE), VisNet-RBF and VisNet-MD achieve **100%** classification accuracy across repeated trials, reflecting the models' ability to extract highly discriminative and transformation-invariant representations in controlled settings.

Table 2 details the network architecture. The LGN+V1 layer processes $32 \times 32$ inputs through 32 Gabor filters, producing $80 \times 80 \times 32$ feature maps. Subsequent layers maintain $80 \times 80$ spatial resolution to balance representational capacity and computational efficiency on RTX 4090 hardware. Table 3 summarizes the classification accuracies obtained across multiple datasets using different VisNet configurations, including the VisNet-Simplified model, VisNet-RBF, and VisNet-MD. The results highlight VisNet's remarkable generalization ability across both synthetic and real-world datasets. Notably, the model achieved a perfect classification accuracy of **100%** on the RGB-IMAGE dataset, underscoring its capacity to capture complex color and symmetry features. Overall, these findings demonstrate the adaptability and robustness of VisNet, reinforcing its potential as a biologically inspired computational model for visual perception. Figure 12 provides a comparative analysis of VisNet-Simplified variants against the original VisNet and the HMAX architecture proposed in Rolls (2015). The plot depicts classification accuracy as a function of the number of training samples per class, illustrating the superior performance of VisNet-LI under low-sample conditions. It also suggests that, with larger training datasets, alternative configurations may achieve comparable or improved performance, reflecting the sensitivity of each model to data volume and learning dynamics.

| Layer Number | Layer Input Size | Layer Output Size |
|---|---|---|
| 0 (LGN+V1) | 32x32 | 80x80x32 |
| 1 | 80x80x32 | 80x80 |
| 2 | 80x80 | 80x80 |
| 3 | 80x80 | 80x80 |
| 4 | 80x80 | 80x80 |

Table 2: Input and output sizes of layers.

# 10 Discussion

## 10.1 Summary of Findings

The results of this study demonstrate that VisNet is effective in classifying and recognizing objects across a range of experimental conditions. The model exhibits strong generalization to unseen data, indicating its ability to capture essential structural properties of visual stimuli through hierarchical processing and Hebbian learning. In particular, the temporal continuity mechanism—which associates multiple views of the same object across time—plays a critical role in achieving invariance to transformations such as scaling, rotation, and structural distortion.

## 10.2 Interpretation of Classification Results

The interpretation of classification performance must take into account both task complexity and class structure. Experiments were conducted using datasets configured for both binary and multi-class classification. In binary classification tasks, an accuracy of 50% corresponds to chance-level performance and therefore reflects poor discriminative ability. In contrast, for multi-class tasks—such as five-class experiments—a 50% accuracy represents a substantial improvement over the 20% baseline expected from random guessing. Across all experimental settings, VisNet and its variants consistently achieved accuracies well above chance level, regardless of the number of classes involved. This pattern indicates that the learned representations are not task-specific artifacts but reflect robust and generalizable feature extraction capabilities.

| **Dataset** | **Method** | **Accuracy ± SD** |
|---|---|---|
| **Standard Datasets** | | |
| MNIST | VisNet-Simplified | 87% ± 2.5% |
| | VisNet-LI | 92% ± 2.0% |
| | VisNet-RBF | 92% ± 1.8% |
| | **VisNet-MD** | **94% ± 2.1%** |
| CIFAR10 | VisNet-Simplified | 25% ± 3.2% |
| | VisNet-LI | 30% ± 2.8% |
| | VisNet-RBF | 35% ± 2.5% |
| | VisNet-MD | 36% ± 2.3% |
| | **VisNet-LI-DoG-RGB-WTA** | **52% ± 2.3%** |
| **Custom Symmetric Datasets** | | |
| RGB-IMAGE | VisNet-Simplified | 94% ± 1.5% |
| | VisNet-LI | 99% ± 0.5% |
| | **VisNet-RBF** | **100% ± 0.0%** |
| | **VisNet-MD** | **100% ± 0.0%** |
| SQUARE | VisNet-Simplified | 38% ± 3.0% |
| | VisNet-LI | 42% ± 2.7% |
| | **VisNet-RBF** | **48% ± 2.4%** |
| | VisNet-MD | 46% ± 2.5% |
| TWOCLASSES-SQUARE | VisNet-Simplified | 80% ± 2.8% |
| | VisNet-LI | 83% ± 2.6% |
| | VisNet-RBF | 88% ± 2.0% |
| | **VisNet-MD** | **92% ± 1.7%** |
| TRIANGLE | VisNet-Simplified | 75% ± 2.9% |
| | VisNet-LI | 78% ± 2.6% |
| | **VisNet-RBF** | **82% ± 2.2%** |
| | VisNet-MD | 81% ± 2.3% |
| ROTATED-TRANSLATED-TRIANGLE | VisNet-Simplified | 62% ± 3.5% |
| | VisNet-LI | 66% ± 3.2% |
| | VisNet-RBF | 74% ± 2.8% |
| | **VisNet-MD** | **82% ± 2.4%** |
| FIVECLASSES-PARTED-SQUARE | VisNet-Simplified | 39% ± 3.1% |
| | VisNet-LI | 42% ± 2.9% |
| | **VisNet-RBF** | **70% ± 2.5%** |
| | VisNet-MD | 67% ± 2.6% |
| TWOCLASSES-PARTED-SQUARE | VisNet-Simplified | 87% ± 2.0% |
| | VisNet-LI | 92% ± 1.8% |
| | **VisNet-RBF** | **100% ± 0.0%** |
| | VisNet-MD | 98% ± 0.8% |
| FIVECLASSES-SOMEPARTED-SQUARE | VisNet-Simplified | 34% ± 3.3% |
| | VisNet-LI | 38% ± 3.0% |
| | **VisNet-RBF** | **44% ± 2.8%** |
| | VisNet-MD | 43% ± 2.9% |
| TWOCLASSES-SOMEPARTED-SQUARE | VisNet-Simplified | 89% ± 1.9% |
| | VisNet-LI | 92% ± 1.7% |
| | **VisNet-RBF** | **100% ± 0.0%** |
| | VisNet-MD | 98% ± 0.9% |
| ROTATED-TRANSLATED-HUMAN-LIKE | VisNet-Simplified | 43% ± 3.2% |
| | VisNet-LI | 45% ± 3.0% |
| | VisNet-RBF | 68% ± 2.7% |
| | **VisNet-MD** | **72% ± 2.5%** |

Table 3: Classification accuracy (± standard deviation) across datasets for VisNet-Simplified and variants.

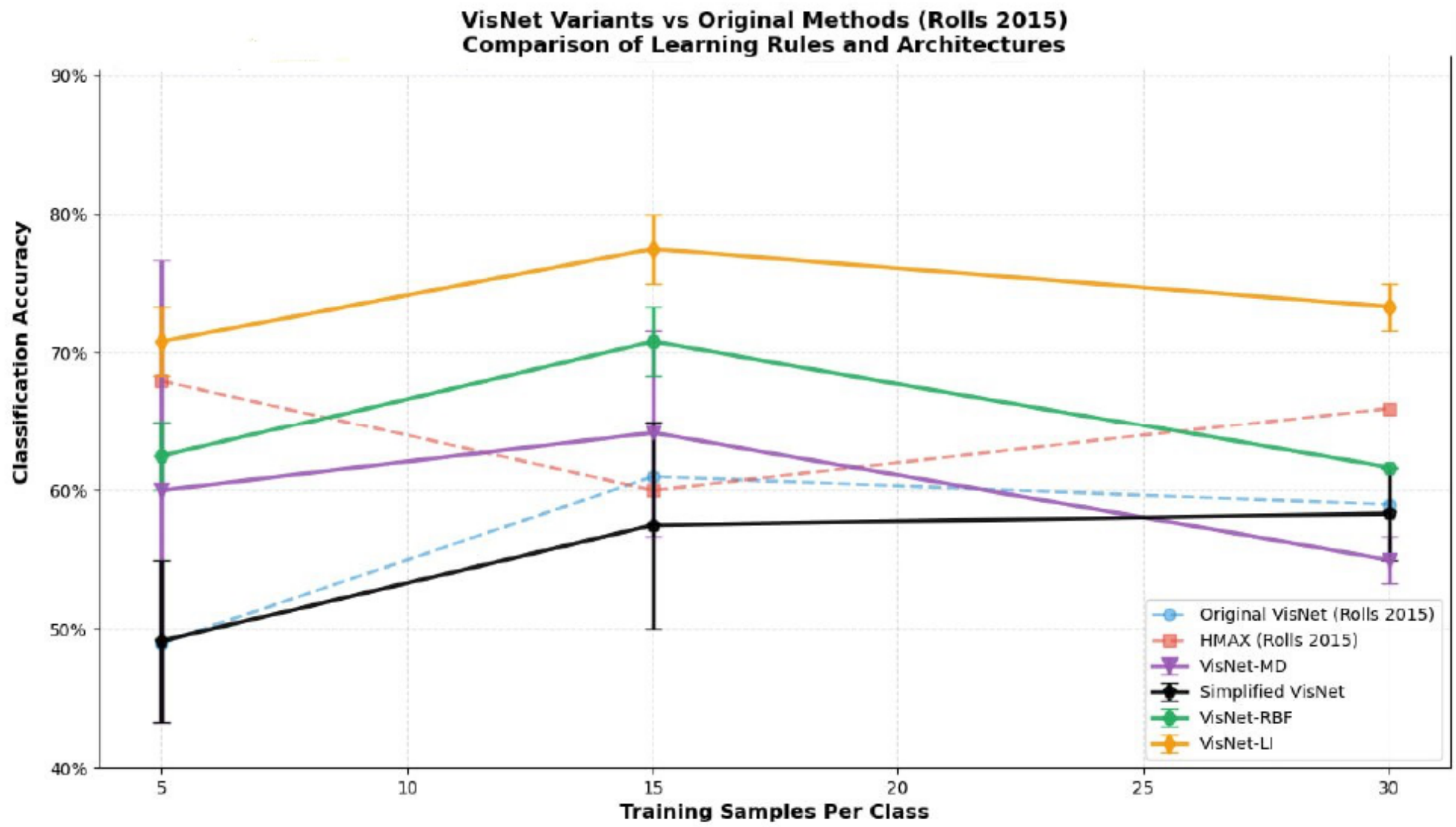

Figure 12: VisNet-Simplified Variants vs Original Methods Rolls (2015). Comparison of Learning Rules and Architectures. Experimental Setup: Unsupervised training on training samples; Linear SVM classification; 30 test samples per class.

## 10.3 Statistical Validity

All reported performance measures were obtained through systematic experimentation. Each experimental condition was repeated multiple times (e.g., $n = 10$ trials) in order to capture variability arising from stochastic factors such as initialization and sampling. We report mean classification accuracy together with its standard deviation:

$$\bar{A} \pm \sigma_A = \frac{1}{n}\sum_{i=1}^{n} A_i \pm \sqrt{\frac{1}{n-1}\sum_{i=1}^{n}(A_i - \bar{A})^2} \tag{13}$$

This evaluation protocol ensures that the reported improvements reflect consistent performance gains rather than random fluctuations. Where appropriate, statistical significance was assessed using paired $t$-tests, with a significance threshold of $p < 0.05$.

## 10.4 Implications for Biological Vision

The present findings carry important implications for understanding biological vision. In human and animal perception, symmetry constitutes a fundamental cue for object recognition and categorization, facilitating the detection of structured and meaningful forms in natural environments Wagemans (1995). The results reported here suggest that computational models such as VisNet can reproduce key aspects of these perceptual processes, providing mechanistic insights into how invariant visual representations may arise in the brain.

The core biological principles instantiated in the VisNet model include:

1. **Hierarchical Processing**: Progressive abstraction of visual features across successive layers, analogous to the organization of the ventral visual stream (V1 $\rightarrow$ V2 $\rightarrow$ V4 $\rightarrow$ IT) DiCarlo et al. (2012b).

2. **Receptive Fields**: Spatially localized and progressively expanding receptive fields that constrain neuronal responses to specific regions of the visual field, enabling the gradual integration of local features into increasingly complex and invariant representations Hubel and Wiesel (1962a); Rolls (2012).

3. **Hebbian Learning**: Local, unsupervised synaptic modification driven by correlations between pre- and post-synaptic activity Hebb (1949b).

4. **Lateral Inhibition**: Competitive interactions that promote sparse, selective, and decorrelated neural representations Rolls (2021b).

5. **Temporal Continuity**: Exploitation of the statistical regularities of natural visual input, whereby consecutive views over time typically correspond to the same object Wallis and Rolls (1997b).

## 10.5 Implications for Artificial Intelligence

From an artificial intelligence perspective, VisNet provides a biologically inspired framework well suited for tasks that require invariant feature recognition under transformations such as changes in viewpoint, position, and scale. By leveraging unsupervised learning principles grounded in neuroscience, VisNet offers an alternative to purely data-driven deep learning approaches. Table 4 summarizes representative application domains where VisNet-based architectures may be particularly effective.

Table 4: Potential Applications of VisNet-Based Architectures

| **Domain** | **Application** |
|---|---|
| Computer Vision | Image classification, object detection |
| Medical Imaging | Symmetry-based anomaly detection |
| Robotics | View-invariant object recognition |
| Biometrics | Face recognition across viewpoints |
| Quality Control | Defect detection in manufactured parts |

Compared to conventional deep learning approaches, VisNet exhibits several distinctive advantages:

- **Unsupervised Feature Formation**: By leveraging Hebbian mechanisms, the framework mitigates "data hunger" by learning robust representations from raw data streams, significantly reducing the reliance on vast, human-annotated datasets.

- **Layer-wise Credit Assignment**: Unlike global backpropagation, each layer optimizes its parameters using local signals. This modularity allows for the training of deep hierarchies without the computational bottleneck of a global loss function, mirroring the modular organization of the brain.

- **Real-time Online Learning**: The system processes and learns from data samples sequentially. This eliminates the need for large memory buffers (batches) and allows the model to adapt continuously to non-stationary environments, a prerequisite for autonomous biological intelligence.

- **Biological Plausibility**: The architecture adheres to the locality principle; learning rules are synapse-specific and rely only on local pre- and post-synaptic activity, bypassing the biological implausibility of "weight transport" found in standard AI.

- **Inherent Interpretability**: The hierarchical features emerge from local competition (lateral inhibition), resulting in sparse representations that map directly to the functional properties of the biological visual cortex.

- **Computational & Energy Efficiency**: By avoiding global error backpropagation and relying on local learning rules, the proposed framework reduces computational complexity and memory access demands, improving energy efficiency on conventional computing architectures. These benefits are particularly well aligned with neuromorphic hardware platforms, such as Intel Loihi Davies et al. (2018), where local, event-driven learning can be exploited for ultra–low-power inference and adaptation.

Despite these advantages, several limitations should be acknowledged:

- Performance on highly complex, large-scale natural image datasets (e.g., ImageNet) currently lags behind state-of-the-art deep learning models.

- The temporal continuity learning paradigm assumes access to structured or sequential input data, which may not be available in all application settings.

- Sensitivity to hyperparameters such as learning rates and inhibitory strengths necessitates careful tuning for stable and optimal performance.

- The absence of attention-like mechanisms limits the model's capacity for sequential data processing, a feature central to transformer-based architectures such as LLMs.

# 11 Conclusion

In this study, we demonstrated that the VisNet model can effectively classify both synthetic symmetric objects and real-world image datasets, including CIFAR-10, by leveraging a set of biologically inspired mechanisms such as hierarchical processing, Hebbian learning, and temporal continuity. Together, these mechanisms enable VisNet to form invariant object representations that maintain recognition performance under challenging transformations, including changes in scale and rotation. Furthermore, the integration of Mahalanobis distance–based learning with radial basis function (RBF) neurons enhances the robustness of the model, allowing it to capture complex data distributions and to support more expressive unsupervised feature learning. Despite these strengths, the relatively modest performance on CIFAR-10 highlights important limitations of the current architecture, particularly in handling the diverse statistical properties of natural RGB imagery. Architectural refinements—such as increasing network depth, expanding the number of neurons and feature channels, and introducing more flexible receptive field organizations—are likely to improve the model's representational capacity. In addition, systematic evaluation on larger and more complex datasets will be necessary to assess the scalability and generalization of VisNet across broader visual domains. Future research will focus on three main directions: (1) optimizing the VisNet architecture to improve performance on real-world visual recognition tasks; (2) extending the framework to quantify and rank objects according to graded levels of symmetry; and (3) applying the model to higher-dimensional datasets to examine robustness in increasingly complex feature spaces. Through these developments, this work contributes toward bridging the gap between biologically inspired models of vision and contemporary artificial intelligence, demonstrating how principles derived from neural computation can inform the design of interpretable, efficient, and biologically grounded AI systems.

# Conflict of Interest Statement

The authors declare that the research was conducted in the absence of any commercial or financial relationships that could be construed as a potential conflict of interest.

# Acknowledgments

This research was conducted at the Computer Vision Center (CVC) of the Autonomous University of Barcelona (UAB) as part of the author's Ph.D. thesis work. The author gratefully acknowledges the support and resources provided by the CVC and UAB throughout the development of this study.

# Data Availability Statement

The code for the VisNet-Simplified model and its variants is publicly available on GitHub at https://github.com/mehdifatan/VisNet.